\documentclass{llncs}
\usepackage{llncsdoc}
\usepackage{graphicx}
\usepackage{xfrac}
\usepackage{verbatim}
\usepackage{amssymb}
\usepackage{amsmath}
\usepackage{float}
\usepackage{booktabs}
\usepackage{upgreek}
\usepackage{hyperref}
\usepackage{xcolor}
\usepackage{bbm}
\usepackage{enumitem}
\usepackage{xcolor}
\usepackage{caption}
\usepackage{subcaption}
\usepackage{multirow}

\newcommand{\beq}{\begin{equation}}
\newcommand{\eeq}{\end{equation}}

\newcommand{\mr}[1]{\mathrm{#1}}

\newcommand{\mypar}[1]{\vspace{.3em}\noindent\textbf{#1}~}

\usepackage{scalerel,stackengine}
\stackMath
\newcommand\reallywidehat[1]{%
\savestack{\tmpbox}{\stretchto{%
 \scaleto{%
 \scalerel*[\widthof{\ensuremath{#1}}]{\kern-.6pt\bigwedge\kern-.6pt}%
 {\rule[-\textheight/2]{1ex}{\textheight}}
 }{\textheight}%
}{0.5ex}}%
\stackon[1pt]{#1}{\tmpbox}%
}

\newcommand{\xb}{\textbf{x}} 
\newcommand{\yb}{\textbf{y}} 

\begin{document}
\frenchspacing


\title{TAAL: Test-time Augmentation for Active Learning in Medical Image Segmentation}


%
\author{M\'elanie Gaillochet\thanks{Corresponding author: M. Gaillochet.   \textbf{Email:}~\email{melanie.gaillochet.1@ens.etsmtl.ca}} \and
Christian Desrosiers \and
Herv\'e Lombaert}

\institute{ETS Montreal, Canada}


\maketitle 

\begin{abstract}
Deep learning methods typically depend on the availability of labeled data, which is expensive and time-consuming to obtain. Active learning addresses such effort by prioritizing which samples are best to annotate in order to maximize the performance of the task model. While frameworks for active learning have been widely explored in the context of classification of natural images, they have been only sparsely used in medical image segmentation. The challenge resides in obtaining an uncertainty measure that reveals the best candidate data for annotation. 
This paper proposes Test-time Augmentation for Active Learning (TAAL), a novel semi-supervised active learning approach for segmentation that exploits the uncertainty information offered by data transformations. Our method applies cross-augmentation consistency during training and inference to both improve model learning in a semi-supervised fashion and identify the most relevant unlabeled samples to annotate next. In addition, our consistency loss uses a modified version of the JSD to further improve model performance. By relying on data transformations rather than on external modules or simple heuristics typically used in uncertainty-based strategies, TAAL emerges as a simple, yet powerful task-agnostic semi-supervised active learning approach applicable to the medical domain. Our results on a publicly-available dataset of cardiac images show that TAAL outperforms existing baseline methods in both fully-supervised and semi-supervised settings. Our implementation is publicly available on https://github.com/melinphd/TAAL.

\end{abstract}

\section{Introduction} 
The performance of deep learning-based models improves as the number of labeled training samples increases. Yet, the burden of annotation limits the amount of data that can be labeled. One solution to that problem is offered by active learning (AL) \cite{settles_active_2009}. Based on the hypothesis that all data samples have a different impact on training, active learning aims to find the best set of candidate samples to annotate in order to maximize the performance of the task model. 
In such context, medical image segmentation emerges as a remarkably relevant task for active learning. Indeed, medical images typically require prior expert knowledge for their analysis and annotation, an expensive and time-consuming task. 
Initial attempts have explored active learning in medical imaging \cite{budd_2021}, but their methodology either relied on simple uncertainty heuristics \cite{top_active_2011,konyushkova_geometry_2019} or required heavy computations during sampling \cite{sourati_intelligent_2019,nath_diminishing_2020} or training \cite{yang_suggestive_2017}.

\mypar{Deep active learning} Active learning has been extensively explored for the classification \cite{ash_deep_2020,beluch_ensembles_2018,gal_bayesian_2017,sener_active_2018,wang_cost-effective_2017,yoo_learning_2019} or segmentation \cite{vezhnevets2012active,siddiqui2020viewal,casanova2019reinforced} of natural images. Recent deep active learning approaches based on entropy \cite{wang_cost-effective_2017} or ensembles \cite{beluch_ensembles_2018} adapted traditional uncertainty-based AL strategies to deep learning models. 
Similarly, DBAL \cite{gal_bayesian_2017} combined measures such as entropy or mutual information with Monte-Carlo dropout to suggest which samples to annotate next. 
Core-set selection \cite{sener_active_2018} aimed to find the best batch sampling strategy for CNNs in classification, but did not scale well to high-dimensional data. 

The use of auxiliary modules \cite{yoo_learning_2019,sinha_variational_2019,kim_task-aware_2021} has been similarly explored to improve AL sampling strategies. The loss prediction module of \cite{yoo_learning_2019} measured model uncertainty with intermediate representations. Likewise, a VAE was used in VAAL \cite{sinha_variational_2019} to learn the latent representation of the unlabeled dataset and distinguish between labeled and unlabeled samples.  While these state-of-the-art methods have improved previous approaches, their dependence on auxiliary modules reduces their flexibility and increase the burden of hyperparameter tuning.

\mypar{Semi-supervised AL} Semi-supervised learning  (SSL) exploits the representations of unlabeled data to improve the performance of the task model. Since semi-supervised learning and active learning are closely connected, recent works in AL have attempted to combine both domains \cite{wang_cost-effective_2017,sinha_variational_2019,kim_task-aware_2021,huang_temporal-discrepancy_2021}. For instance, CEAL \cite{wang_cost-effective_2017} used pseudo-labeling of unlabeled samples to enhance the labeled set during training. VAAL \cite{sinha_variational_2019} and TA-VAAL \cite{kim_task-aware_2021} employed a VAE to learn a latent representation of labeled and unlabeled data. The Mean Teacher framework of \cite{huang_temporal-discrepancy_2021} combined a supervised loss on labeled data with an unsupervised loss on unlabeled data based on Temporal Output Discrepancy (TOD), evaluating the distance between the model's output at different gradient steps. The model used a variant of TOD at sampling time to identify the most uncertain samples to annotate. However, these semi-supervised AL methods solely focused on classification tasks or the segmentation of natural images in very large quantities, which is a different context than medical imaging. Another recent work  comparable to ours combined AL and SSL via consistency regularization \cite{gao_consistency-based_2020}. The consistency loss adopted during training employed MixMatch \cite{berthelot_mixmatch_2019} and sample selection measured inconsistency across input perturbations. However, as opposed to our work, \cite{gao_consistency-based_2020} kept the consistency loss used during training and the AL inconsistency metric used for sample selection independent of each other, and the latter was quantified through variance. Furthermore, the method was only validated on classification tasks.

\mypar{Test-time augmentation} Data augmentation is a well-known regularization technique to improve generalization in low-data regimes. These augmentation techniques are particularly essential in medical imaging where datasets tend to be smaller than those of natural images. Yet most recent attempts in active learning do not exploit data augmentation during training \cite{ash_deep_2020,nath_diminishing_2020}, or only use random horizontal flipping \cite{sinha_variational_2019,kim_task-aware_2021}. 
Recent learning methods \cite{ayhan_test-time_2018,wang_TestTimeAUg_2019} 
have also investigated the use of augmentation at test-time in order evaluate prediction uncertainty. Randomly augmented test images yield different model outputs. 
Combining these outputs can improve the overall predictions as well as generate uncertainty maps for these predictions. Uncertainty estimated through test-time augmentation was shown to be more reliable than model uncertainty measures such as test-time dropout or entropy of the output \cite{wang_TestTimeAUg_2019}. 

Motivated by the limitations of current active learning methods for medical image segmentation and the unused potential of active augmentation, this paper proposes a novel semi-supervised active learning strategy called Test-time Augmentation for Active Learning (TAAL).

\mypar{Our contribution:} Our method leverages the uncertainty information provided by data augmentation during both training and test-time sample selection phases.
More specifically, TAAL employs a cross-augmentation consistency loss both to train the model in a semi-supervised fashion \emph{as well as} to identify the most uncertain samples to annotate at the next cycle.
TAAL comprises three key features:
\begin{enumerate}
    \item a semi-supervised framework based on cross-augmentation consistency that exploits unlabeled samples during training and sampling;
    \item a flexible task-agnostic sample selection strategy based on test-time augmentation;
    \item a novel uncertainty measure based on a modified Jensen-Shannon divergence (JSD), which accounts for both cross-augmentation consistency and prediction entropy, and leads to improved performance.
\end{enumerate}

\section{Method} 
\mypar{Cross-augmentation consistency training}
We consider a semi-supervised setting where we train a multi-class segmentation model $f_{\theta}(\cdot)$ parameterized by $\theta$ with $N$ labeled samples and $M$ unlabeled samples. We denote the labeled set as $\mathcal{D}_L = \{ (\xb^{(j)}, \yb^{(j)}) \}_{j = 1}^{N}$ and the unlabeled set as $\mathcal{D}_U = \{ \xb_u ^{(j)} \}_{j = 1}^{M}$, with data $\xb, \xb_u \in \mathbb{R}^{H \times W}$ and segmentation mask $\yb \in \mathbb{R}^{C \times H \times W}$ ($C$ is the number of classes).

The overall loss that we optimize,  $\mathcal{L} = \mathcal{L}_s \,+\, \lambda \mathcal{L}_c$, is a combination of a supervised segmentation loss $\mathcal{L}_s$ and an unsupervised consistency loss $\mathcal{L}_c$ weighted by a factor $\lambda$.
More explicitly, the objective is defined as
\begin{equation}
\mathcal{L} \, =  \, \frac{1}{N}\sum_{j = 1}^{N} \mathcal{L}_s \big(f_\theta(\xb^{(j)}), \yb^{(j)}\big) \, + \, \frac{\lambda}{M} \sum_{j = 1}^{M} \mathcal{L}_c  \big(f_\theta(\xb_u^{(j)}), \Gamma\big),
\end{equation}
where $\Gamma$ are the transformations applied to $\xb_u^{(j)}$. At each iteration, we apply a series of random transformations $\{ \Gamma_1, ..., \Gamma_K \}$ to $\xb_u$. $\mathcal{L}_c$ measures the variability of segmentation predictions for different augmentations of $\xb_u$ measured by a function $\mathcal{D}iv$:
\begin{equation}
\mathcal{L}_c \big(f_\theta(\xb_u^{(j)}), \Gamma\big) \, = \, \mathcal{D}iv  \big\{
\Gamma_1^{-1}[f_\theta(\Gamma_1(\xb_u^{(j)}))],\, ...\,, \Gamma_K^{-1}[f_\theta(\Gamma_K(\xb_u^{(j)}))]\big\}.
\end{equation}

While different measures can be used for $\mathcal{D}iv$~\cite{camarasa2020quantitative}, our consistency loss builds on the Jensen Shannon divergence (JSD),
\begin{equation}
\mr{JSD}(P_1, ..., P_K) \, = \,  H\big(\frac{1}{K} \sum_{i = i}^{K}  P_i\big) \, - \,  \frac{1}{K} \sum_{i = i}^{K} H(P_i),
\end{equation}
where $H(P_i)$ is the Shannon entropy \cite{shannon_mathematical_1948} for the probability distributions $P_i$. Minimizing the JSD reduces the entropy of the average prediction (making the predictions more similar to each other) while increasing the average of individual prediction entropies (ensuring confident predictions). In AL we typically want to select samples which have a high output entropy~\cite{wang_cost-effective_2017}. Selecting samples with highest JSD would thus have the opposite effect. To avoid this issue, and to control the relative importance of average prediction entropy versus entropy of individual predictions, we propose a weighted version of JSD with parameter $\alpha$.
\begin{equation}
\mr{JSD}_{\alpha}(P_1, ..., P_K) \, = \, \alpha H\big(\frac{1}{K} \sum_{i = i}^{K}  P_i\big) \, - \,  \frac{(1\!-\!\alpha)}{K} \sum_{i = i}^{K} H(P_i).
\end{equation}

 Note that using $\alpha=0.5$ is equivalent to using the standard JSD.

\mypar{Test-time augmentation sampling}
In active learning, the goal is to select the best unlabeled samples to annotate after each training cycle to augment the next labeled training set. Hence, after each cycle, we apply our active learning strategy based on test-time augmentation to select the next samples to annotate. 

For each sample $\xb_u \in \mathcal{D}_U$, we apply a series of transformations $\{ \Gamma'_1, \ldots, \Gamma'_{K_s} \}$, and we compute an uncertainty score $U_{\Gamma'}$ based on the same divergence function as the consistency loss:
\begin{equation}
U_{\Gamma'} \, = \, \mr{JSD}_{\alpha} \big(
\Gamma_1^{\prime-1}[f_\theta(\Gamma'_1(\xb_u))],\, ...\,, \Gamma_{K_s}^{\prime-1}[f_\theta(\Gamma'_{K_s}(\xb_u))]\big).
\end{equation}

The samples with highest uncertainty are annotated and added to the labeled training set. After sample selection, the model goes through a new training cycle.

\section{Experiments and results}

\subsection{Implementation details}

\subsubsection{Dataset} 
The publicly available ACDC dataset \cite{bernard_acdc_2018} comprises cardiac 3D cine-MRI scans from 100 patients. These are evenly distributed into 5 groups (4 pathological and 1 healthy subjects groups).

Segmentation masks identify 4 regions of interest: right-ventricule cavity, left-ventricule cavity, myocardium and background. For comparative purposes, our experiments focus on the MRI scans at the end of diastole. Preprocessing of the volumes includes resampling to a fixed 1.0\,mm\,$\times$\,1.0\,mm resolution in the x- and y-directions as well as a $99^{th}$ percentile normalization. 
The 3-dimensional dataset of volumes are converted to a 2-dimensional dataset of images by extracting all the z-axis slices for each volume. Each image is downsampled to 128\,$\times$\,128 pixels. Testing is performed on 181 images taken from 20 different patients, ensuring subjects are not split up across training and testing sets. 
The validation uses 100 randomly selected images. The same validation set is used for all experiments. 
In total, the available training set, both labeled and unlabeled, thus comprises 660 images.

\subsubsection{Implementation and training}
We employ a standard 4-layer UNet \cite{ronneberger_u-net_2015} for our backbone segmentation model with dropout ($\mr{p}=0.5$), batch normalization and a leaky ReLU activation function. For a fairer comparison in our experiments, we keep the number of training steps fixed during all cycles. We train our models for 75 epochs, each iterating over 250 batches, with $BS = 4$. We use the Adam optimizer \cite{Adam}, with $LR = 10^{-6}$ and weight decay $w = 10^{-4}$. To improve convergence, we apply a gradual warmup with a cosine annealing scheduler \cite{loshchilov_sgdr_2017,goyal_accurate_2018}, increasing the learning rate by a factor 200 during the first 10 epochs. During training, we apply data augmentation, using transformations similar to those utilized for the consistency loss.

In this work, we model the  transformations $\Gamma$ as a combination of $f$, $r$ and $\epsilon$, where $f$ is the random variable for flipping the image along the horizontal axis, $r$ is the number of $90^{\circ}$ rotations in 2D, and $\epsilon$ models Gaussian noise. We set $f \sim \mathcal{U}(0, 1)$, $r \sim \mathcal{U}(0, 3)$ and $\epsilon \sim \mathcal{N}(0, 0.01)$, and use $K=3$ transformations to compute the consistency loss during training. 

We use the standard Dice loss as our supervised loss. In the semi-supervised case, following \cite{cui_semi-supervised_2019}, we ramp-up the unsupervised component weight using a Gaussian ramp-up curve such that $\lambda = \exp(-5 (1 - t/t_{R})^2)$, where $t$ is the current epoch. We use a ramp-up length $t_{R}$ of 10 epochs, corresponding to the learning rate gradual warmup length.

We repeat each experiment 5 times, each with a different seed determining different initialization of our model weights. For all experiments, the same initial labeled set is used for the first cycle. Experiments were run on NVIDIA PV100 GPU with CUDA 10.2 and Python 3.8.10. We implemented the methods using the PyTorch framework.

\subsubsection{Evaluation metrics}
To evaluate the performance of the trained models, we employ the standard Dice similarity score, averaged over all non-background channels. We compute both the mean 3D Dice on test volumes and mean 2D Dice on the individual images from these volumes. We give the results as the mean Dice obtained over the repeated experiments.

\subsection{Active learning setup}
We begin each experiment with 10 labeled samples chosen uniformly at random in the training set and use a sampling budget of 1, meaning that we select one new sample to be labeled after each cycle. Following previous active learning validation settings~\cite{sener_active_2018}, we retrain the model from scratch after each annotation cycle.
We use the same types of augmentations during training and sample selection. For test-time augmentation (TTA) sampling, $\{ \Gamma'_1, \ldots, \Gamma'_{K_s} \}$ comprises all 8 combinations of flip and rotation augmentations, in order to apply similar transformations to all images, and adopts the same augmentation Gaussian noise parameters as for training. For comparative purposes, with dropout-based sampling, we also run 8 inferences with dropout to obtain different predictions. Both TTA and dropout-based sampling then evaluate uncertainty with $U_{\Gamma'}$ computed on the different generated predictions. We set $\alpha = 0.75$ in TAAL's weighted JSD.

\subsection{Comparison of active learning strategies}
Our aim is to evaluate the effectiveness of our proposed semi-supervised active learning approach on a medical image segmentation task. In our active learning experiments, we compare TAAL and its unweighted version (with standard JSD) with random sampling, entropy sampling, sampling based on dropout and core-set selection. Entropy-based sampling selects the most uncertain samples based on the entropy of the output probabilities. Dropout-based sampling \cite{gal_bayesian_2017} identifies the samples with the highest JSD given multiple inferences with dropout. Finally, core-set selection \cite{sener_active_2018} aims to obtain the most diverse labeled set by solving the maximum cover-set problem.
\vspace{-1em}

\begin{figure}[H]
    \centering
    \hspace*{-3ex}
    \includegraphics[width=0.95\textwidth]{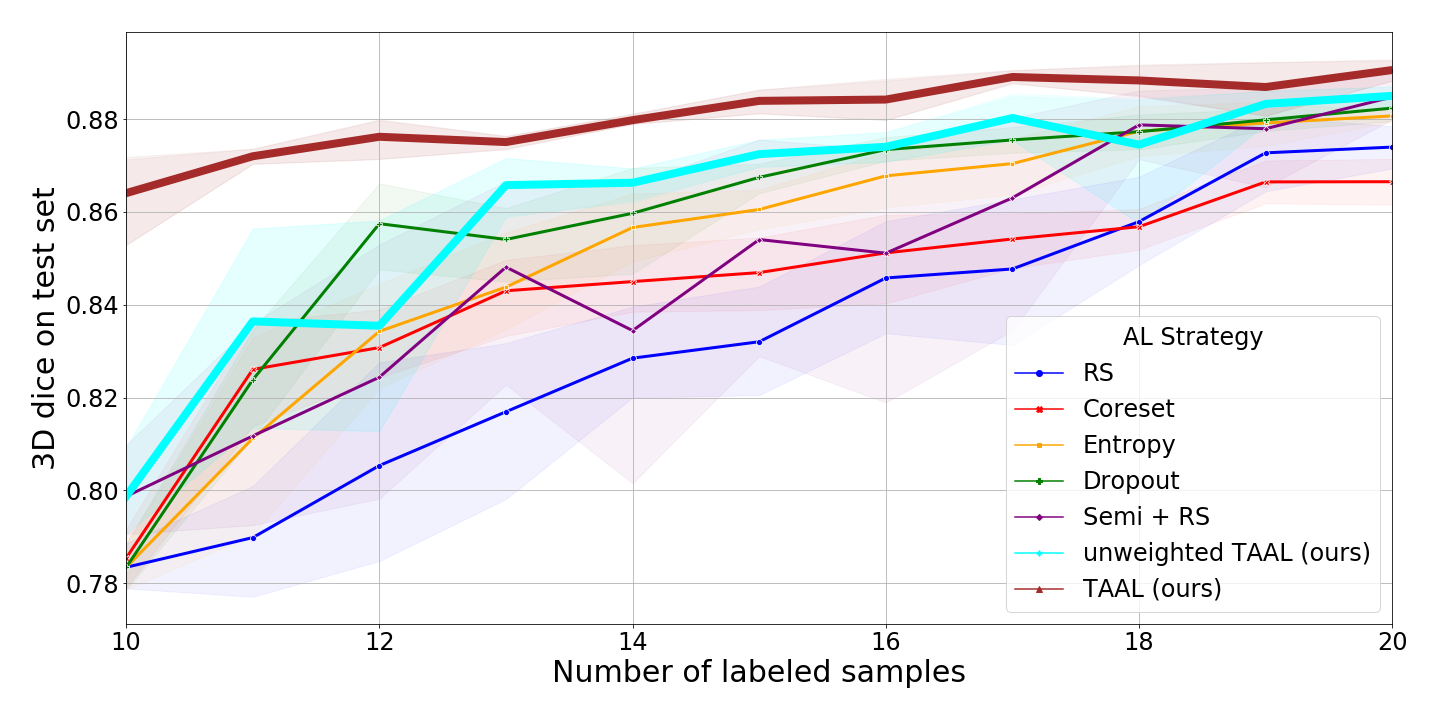}
    \caption{Active learning results on the ACDC dataset, given as the mean 3D Dice scores on the test set and corresponding 95\% confidence interval. In a fully-supervised setting: random sampling (RS), core-set selection (Coreset), uncertainty-based sampling based on entropy of output probabilities (Entropy), and uncertainty-based sampling based on JSD given multiple inferences with dropout (Dropout). In a semi-supervised setting: random sampling (Semi\,+\,RS), TAAL with standard JSD (unweighted TAAL), and TAAL with weighted JSD (TAAL). Our approach TAAL demonstrates significant improvements for low-data regimes in both fully and semi-supervised segmentation.}
    \label{fig:results_TAAL}
\end{figure}
\vspace{-1em}

Figure~\ref{fig:results_TAAL} shows the segmentation performance of our proposed method with its 2 variants along with other existing active learning methods. TAAL consistently outperforms the other baselines by a large margin. We observe that our semi-supervised approach based on cross-augmentation consistency (Semi\,+\,RS) noticeably improves the fully-supervised vanilla model (RS). We notice that our unweighted version of TAAL (with standard JSD, $\alpha\! =\! 0.5$) already improves the performance of the semi-supervised model (Semi\,+\,RS) by selecting the most uncertain samples based on their cross-augmentation consistency loss. With higher $\alpha\! =\! 0.75$, our proposed TAAL with weighted JSD yields the highest performance gain compared to the fully-supervised vanilla model with random sampling (RS).
\vspace{-1em}

\begin{figure}[H]
    \centering
    \begin{subfigure}{0.21\textwidth}
        \includegraphics[width=\linewidth]{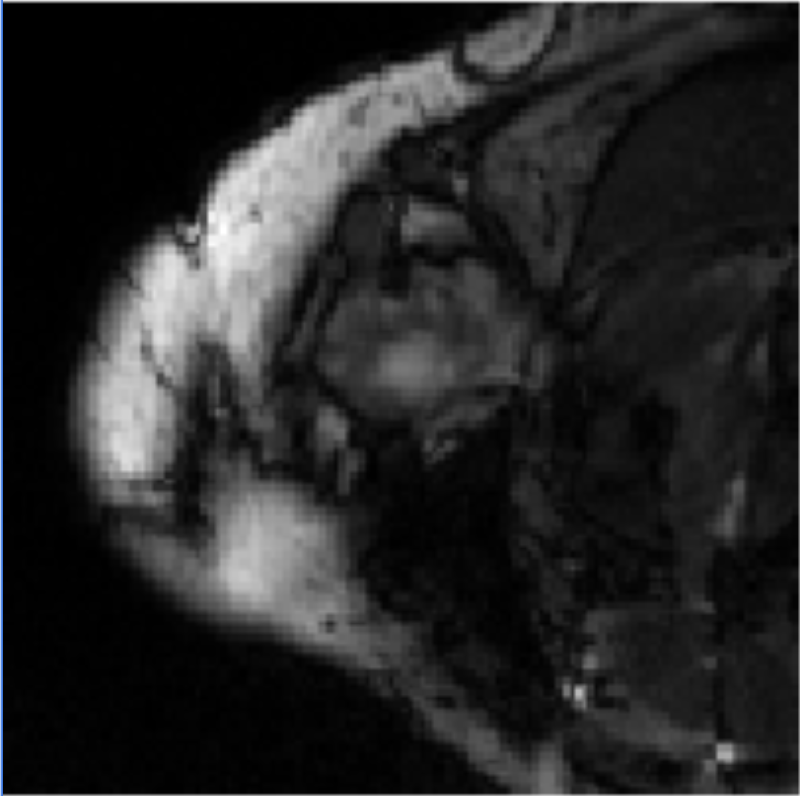}
    \end{subfigure}\hfil 
    \hfil
    \begin{subfigure}{0.21\textwidth}
      \includegraphics[width=\linewidth]{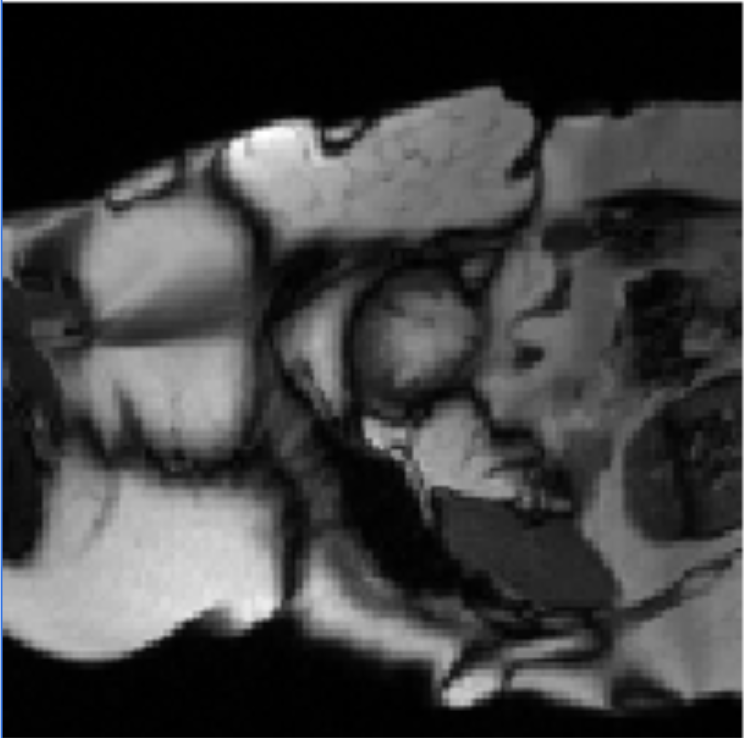}
    \end{subfigure}\hfil 
    \hfil
    \begin{subfigure}{0.21\textwidth}
      \includegraphics[width=\linewidth]{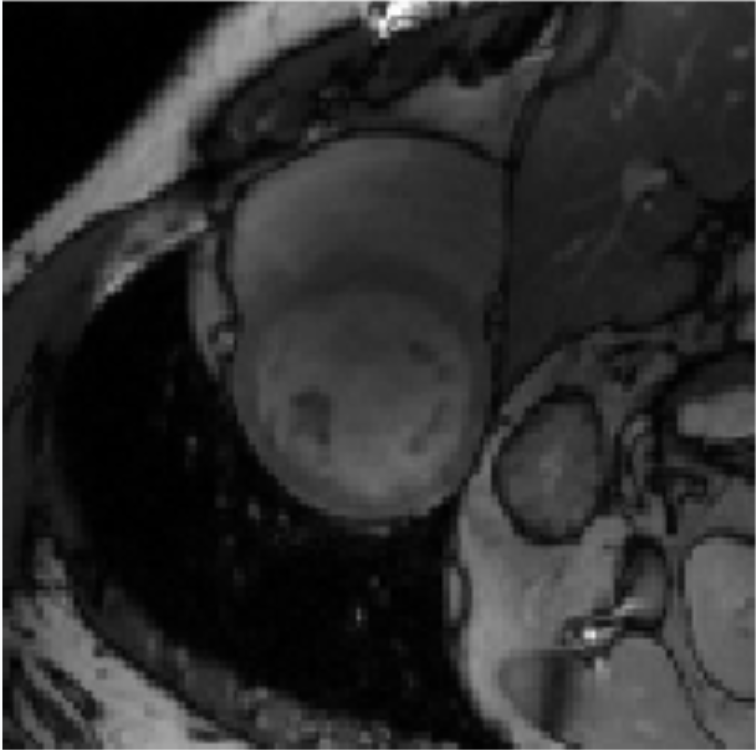}
    \end{subfigure}\hfil 
    \hfil
    \begin{subfigure}{0.21\textwidth}
      \includegraphics[width=\linewidth]{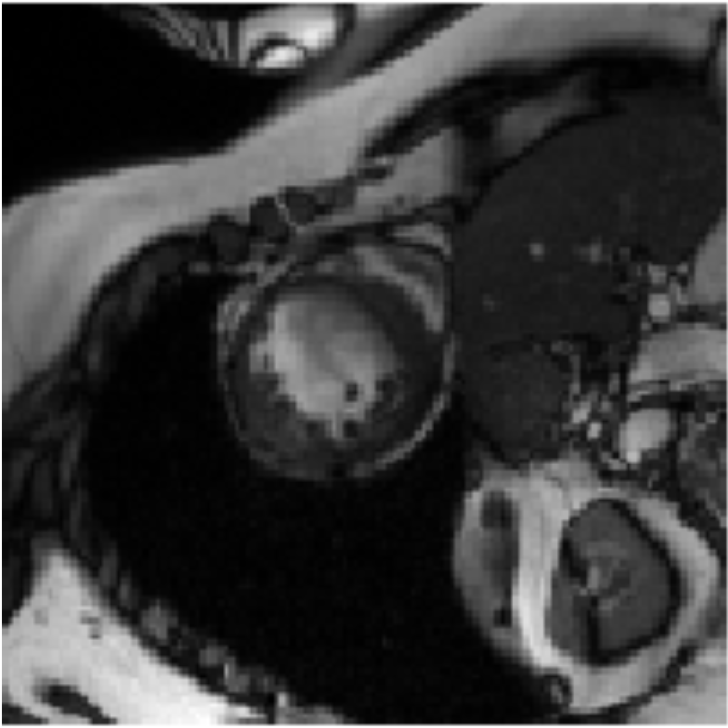}
    \end{subfigure} \hfil 
    \medskip 
    
    \begin{subfigure}{0.21\textwidth}
        \includegraphics[width=\linewidth]{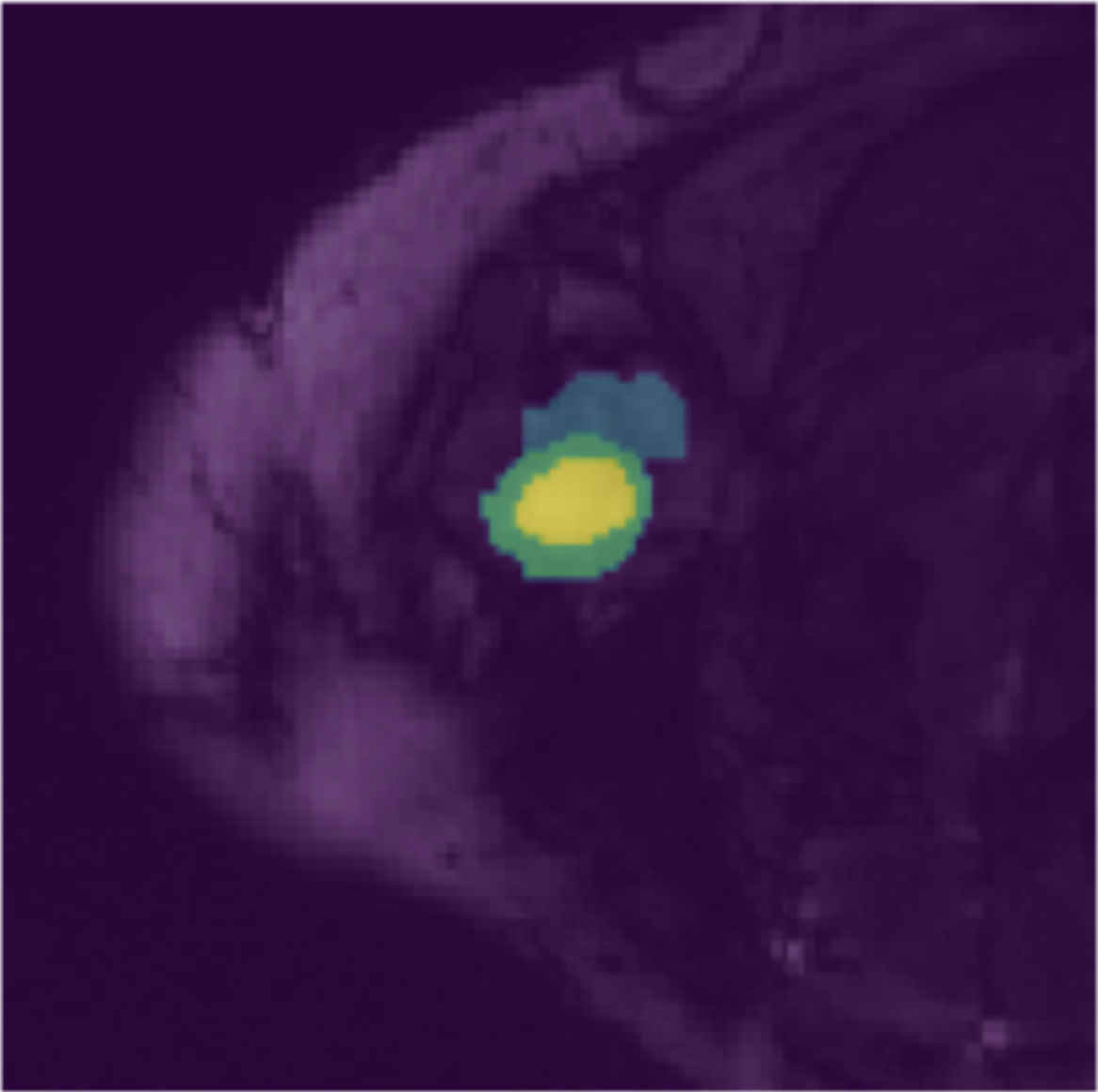}
    \end{subfigure}\hfil 
    \hfil
    \begin{subfigure}{0.21\textwidth}
      \includegraphics[width=\linewidth]{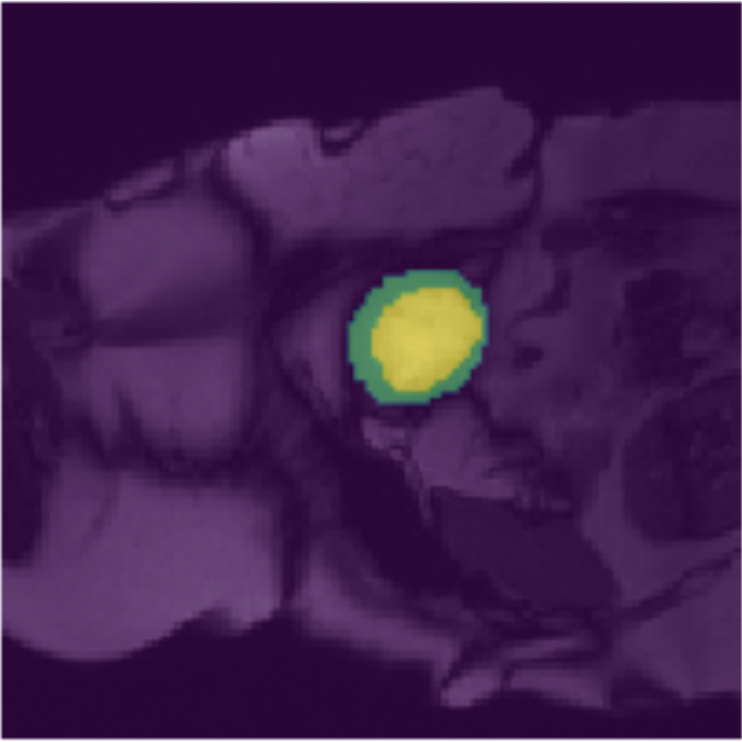}
    \end{subfigure}\hfil 
    \hfil
    \begin{subfigure}{0.21\textwidth}
      \includegraphics[width=\linewidth]{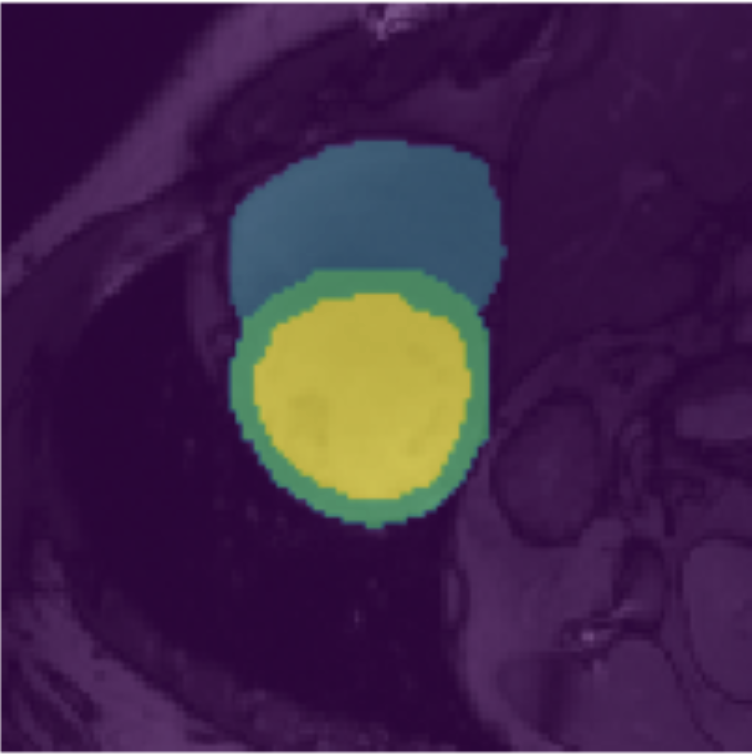}
    \end{subfigure}\hfil 
    \hfil
    \begin{subfigure}{0.21\textwidth}
      \includegraphics[width=\linewidth]{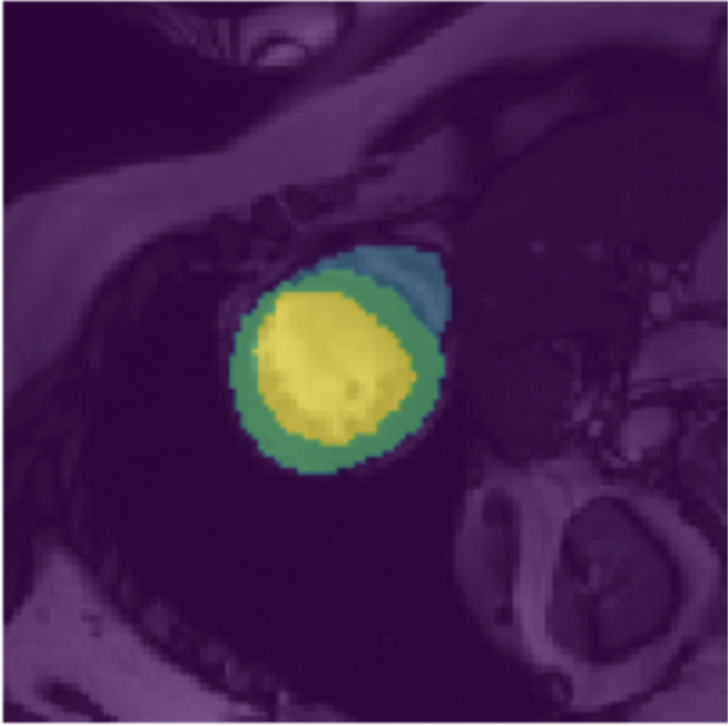}
    \end{subfigure} \hfil 
    \medskip 
    
    \begin{subfigure}{0.21\textwidth}
        \includegraphics[width=\linewidth]{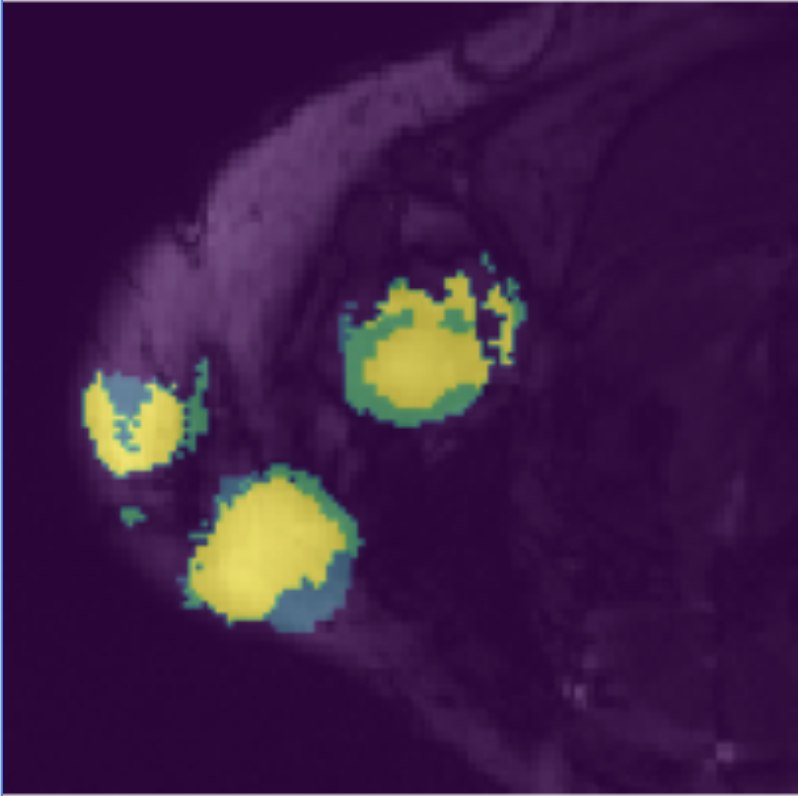}
    \end{subfigure}\hfil 
    \hfil
    \begin{subfigure}{0.21\textwidth}
      \includegraphics[width=\linewidth]{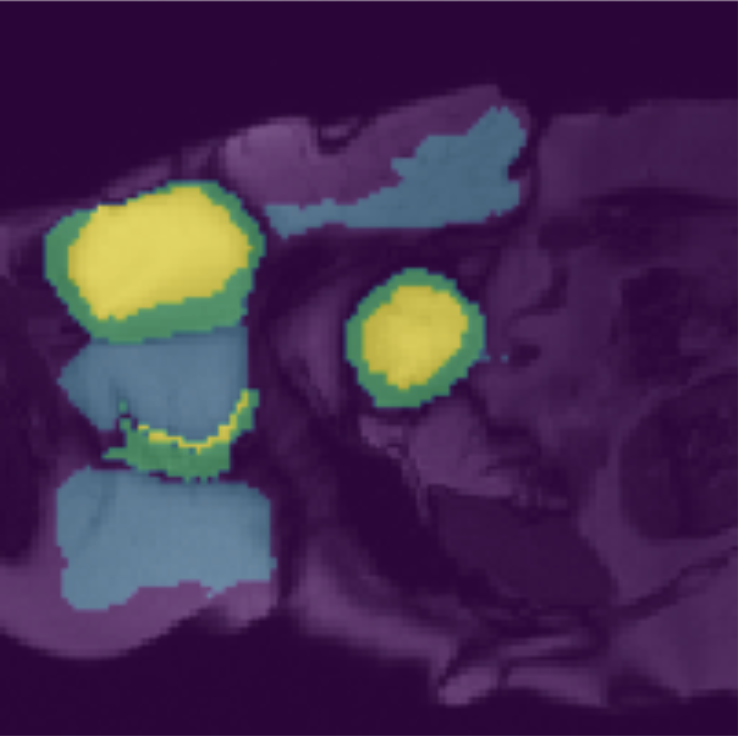}
    \end{subfigure}\hfil 
    \hfil
    \begin{subfigure}{0.21\textwidth}
      \includegraphics[width=\linewidth]{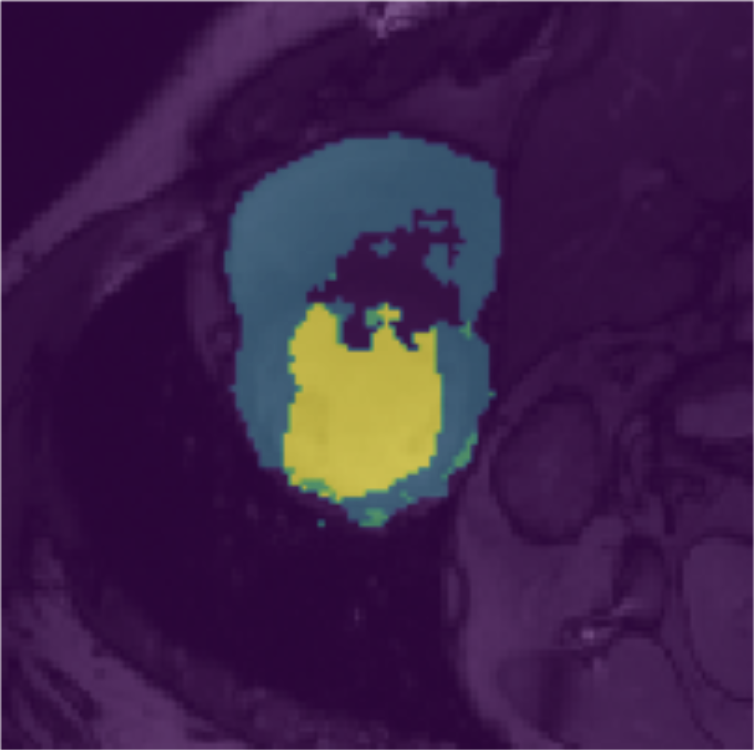}
    \end{subfigure}\hfil 
    \hfil
    \begin{subfigure}{0.21\textwidth}
      \includegraphics[width=\linewidth]{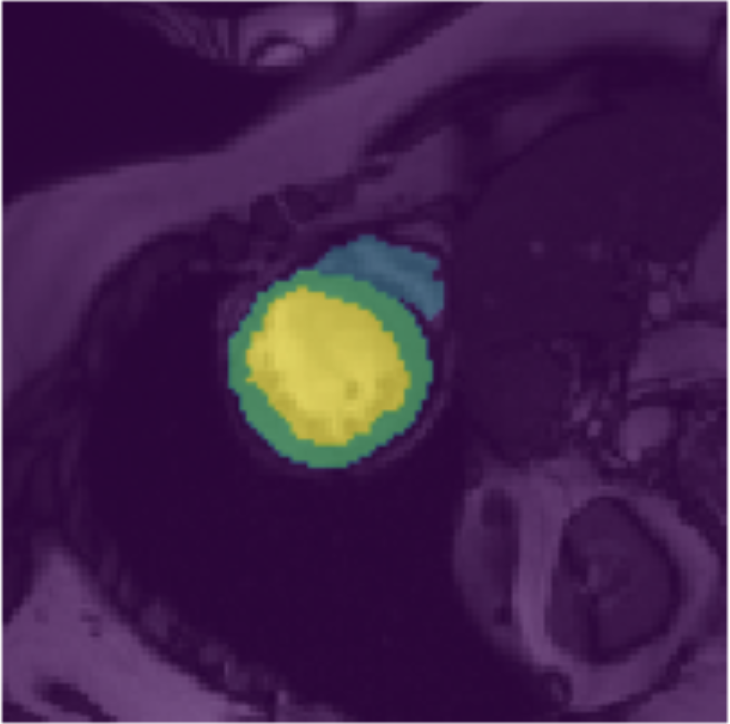}
    \end{subfigure} \hfil 
    \medskip 
    
    \begin{subfigure}{0.21\textwidth}
        \includegraphics[width=\linewidth]{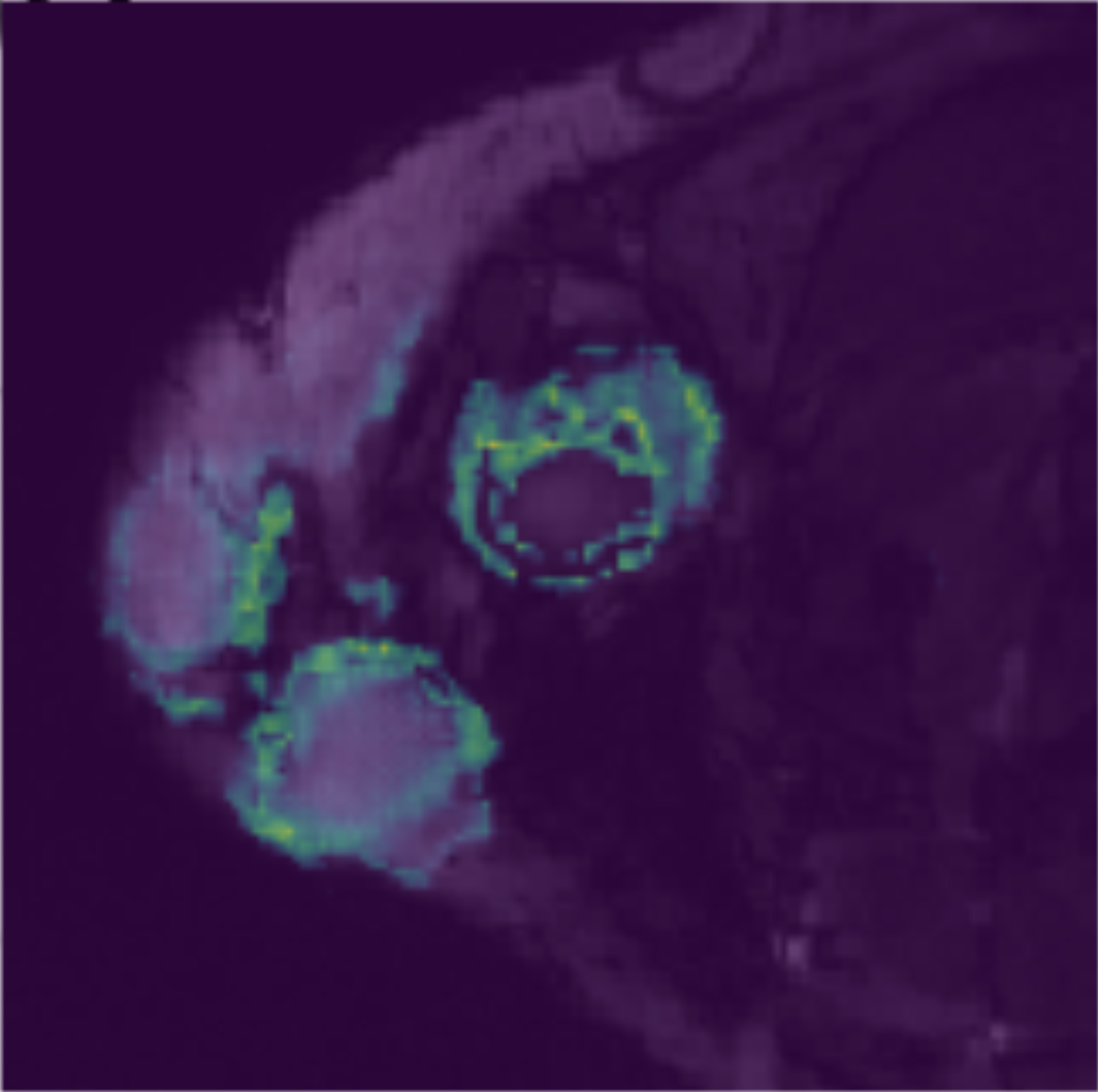}
        \caption{1st Cycle\label{fig:sub:cycle1}}
    \end{subfigure}\hfil 
    \hfil
    \begin{subfigure}{0.21\textwidth}
      \includegraphics[width=\linewidth]{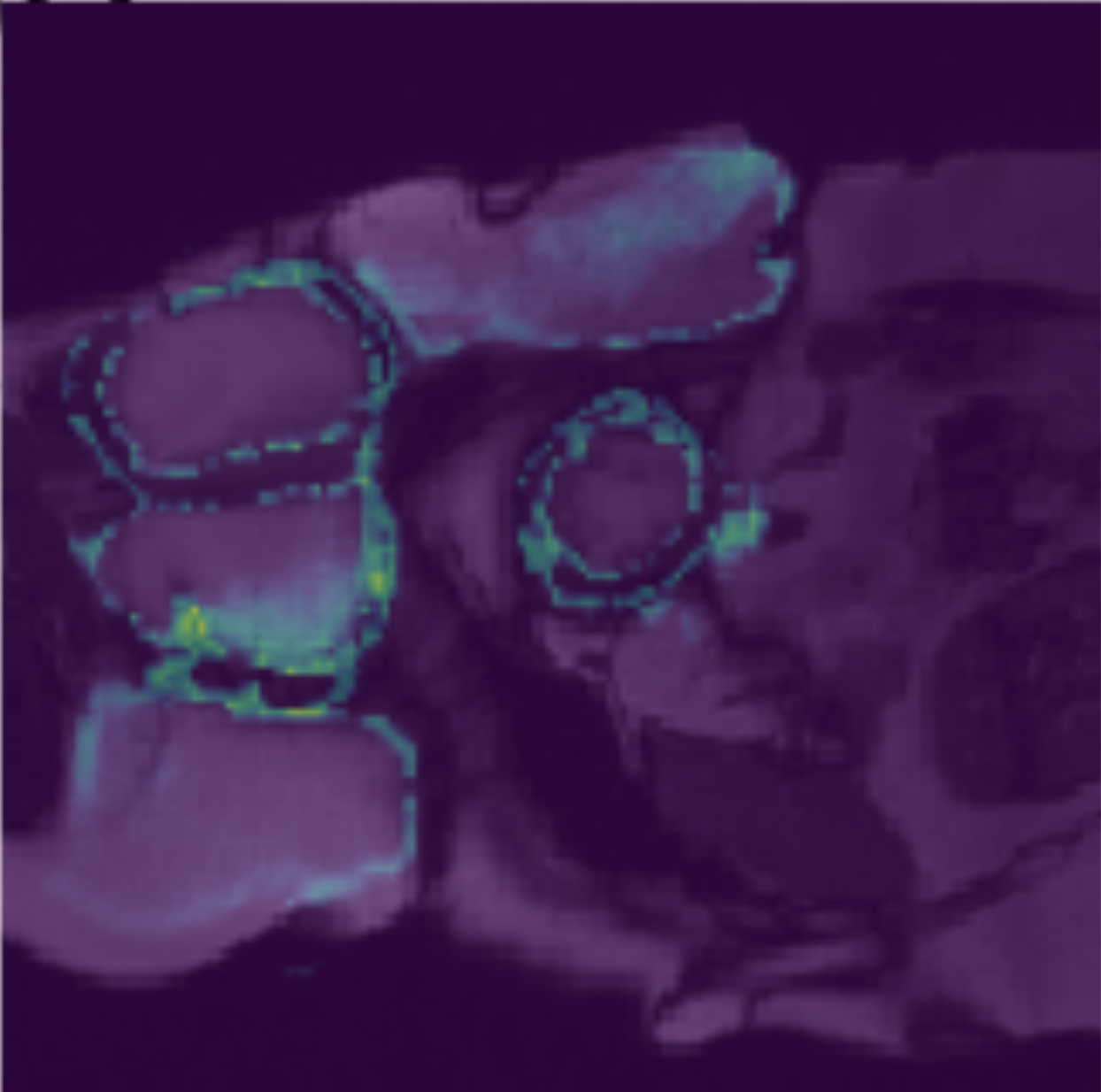}
    \caption{2nd Cycle \label{fig:sub:cycle2}}
    \end{subfigure}\hfil 
    \hfil
    \begin{subfigure}{0.21\textwidth}
      \includegraphics[width=\linewidth]{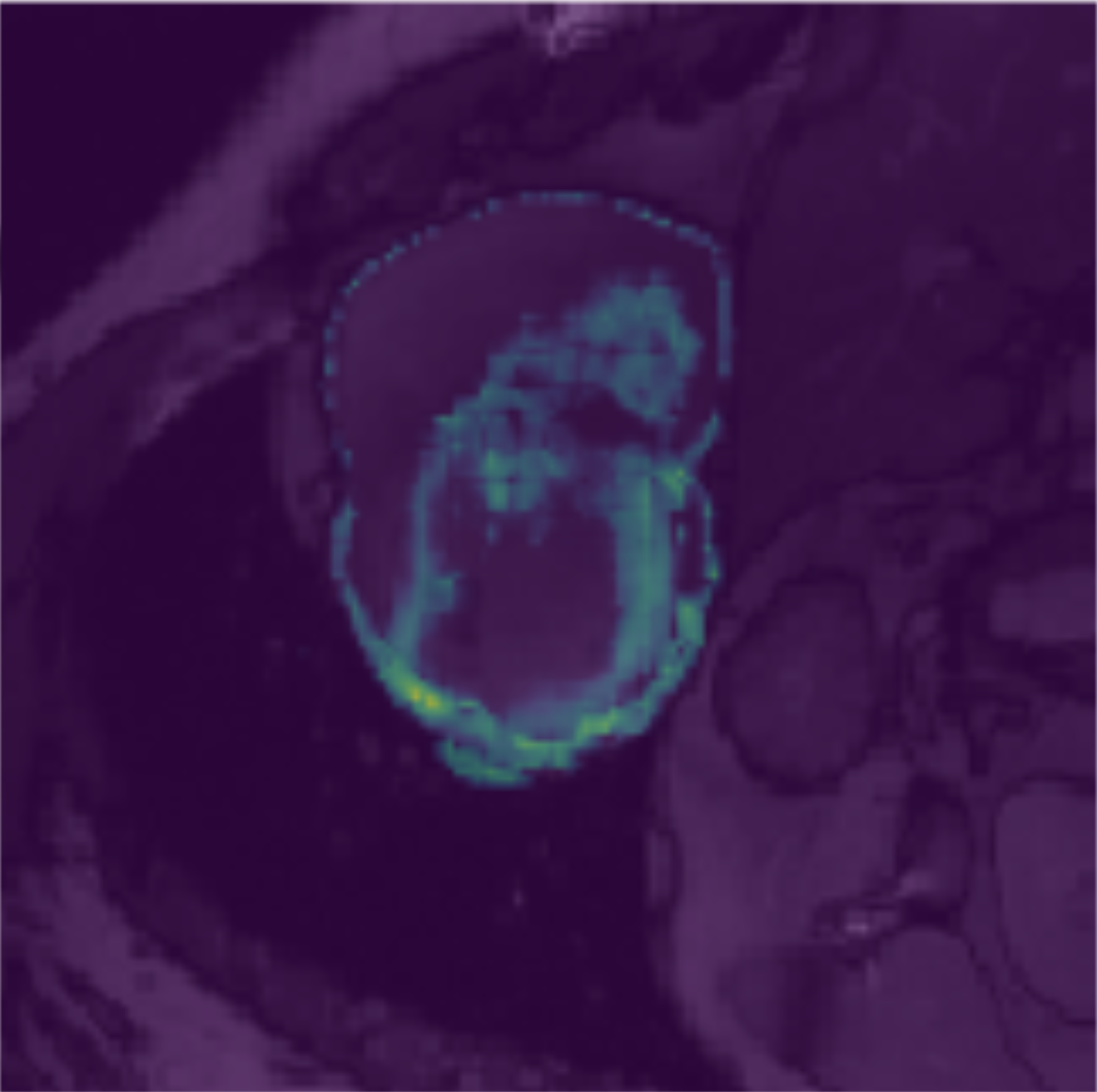}
      \caption{3rd Cycle \label{fig:sub:cycle3}}
    \end{subfigure}\hfil 
    \hfil
    \begin{subfigure}{0.21\textwidth}
      \includegraphics[width=\linewidth]{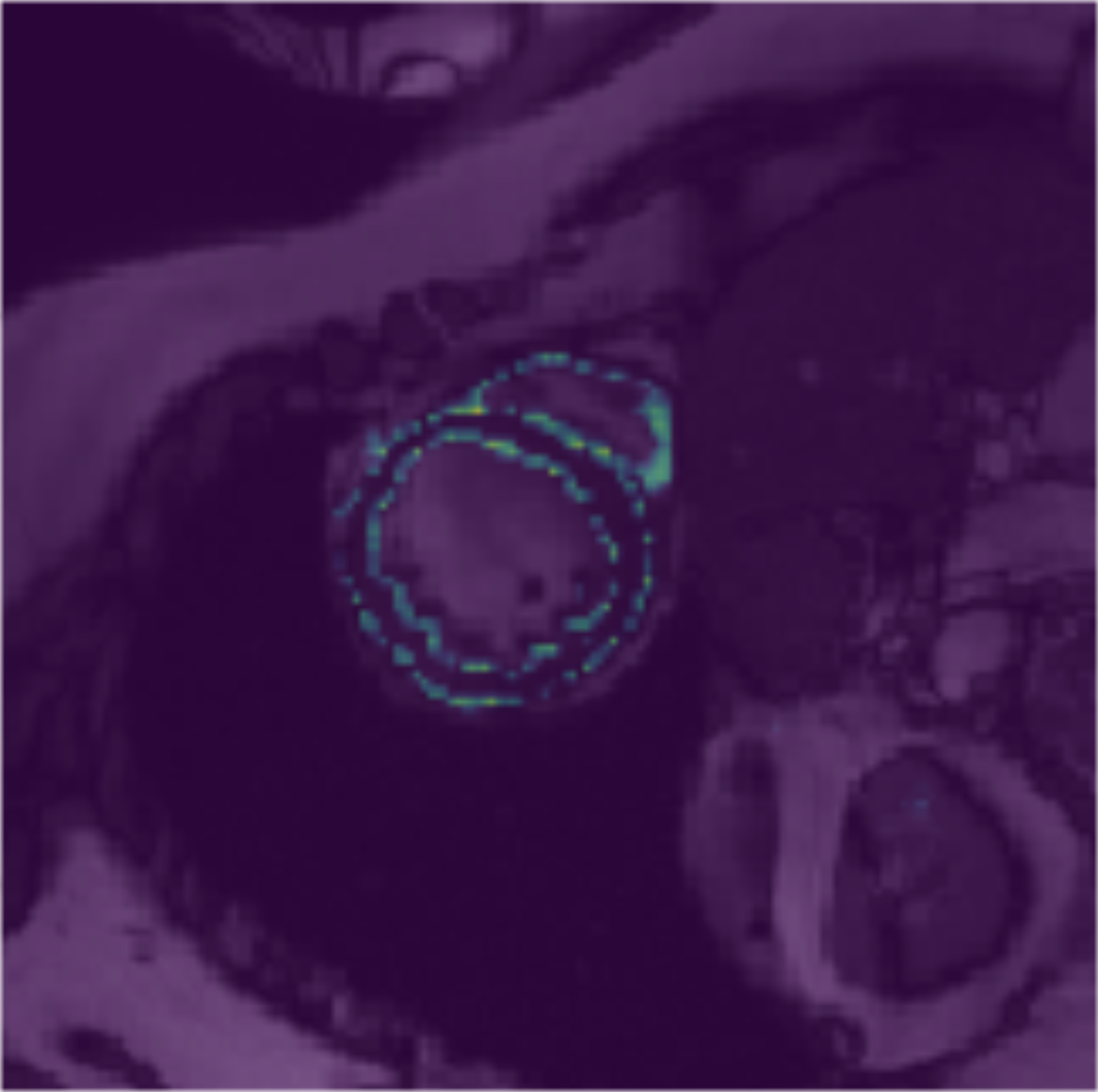}
      \caption{4th Cycle \label{fig:sub:cycle4}}
    \end{subfigure} \hfil 
    \medskip 
    \caption{Examples of images sampled by TAAL at different AL cycles. Are depicted the image sampled (row 1), the ground-truth segmentation (row 2), the segmentation prediction (row 3), and the JSD map given the different predictions from the augmented image (row 4). 
    We observe that TAAL initially selected images with a large amount of hallucinated inaccurate predictions.
    }
    \label{fig:examples}
\end{figure}
\vspace{-1em}

Figure~\ref{fig:examples} shows examples of images sampled by TAAL during the first 4 annotation cycles. TAAL initially selects image slices which show the apex of the heart. These samples are more difficult to learn in early stages since the areas to segment are much smaller than in the central slices of the heart and the image qualities are typically of lesser quality due to partial volume effects. Thus, we see that the choice of TAAL is first directed at samples yielding highly inaccurate predictions. The previous model has in fact even hallucinated multiple false segmentations for these samples as seen on the third row of subfigures~\ref{fig:sub:cycle1} and \ref{fig:sub:cycle2}. In the next cycles, TAAL selects more central cardiac slices, which have improved predictions when compared to the ground-truth annotations. 
Hence, TAAL seems to first focus on correcting inaccurate predictions, before sharpening its predictions on a fine-grained level for slices with more prominent areas to segment.
\vspace{-1.5em}

\begin{table} [H]
    \centering
    \caption{Active learning performances after doubling the number of initial labeled samples. We show the mean 2D and mean 3D Dice scores. `Fully': Fully-supervised vanilla UNet. `Semi': Proposed semi-supervised training with standard ($\alpha=0.5$) or weighted ($\alpha=0.75$) JSD. `RS': Random sampling. `TTA': Sampling with Test-time augmentation. `unweighted TAAL': Our proposed method with standard JSD. `TAAL': Our proposed method with weighted JSD, which finds the best candidate image to annotate.}
    \vspace{1.0em}
    \resizebox{\textwidth}{!}{
    \setlength{\tabcolsep}{4pt}
    \begin{tabular}{ c | c | c  | c | c | c || c | c | c}
        \toprule
         \multirow{2}{*}{Metric} & \multicolumn{5}{c||}{Fully} & \multicolumn{2}{|c|}{Semi ($\alpha=0.5$)} & Semi ($\alpha=0.75$)\\
         \cmidrule(l{4pt}r{4pt}){2-9}
           & RS & Coreset & Entropy & Dropout  & TTA & RS &  unweighted TAAL & TAAL\\
        \midrule
        2D Dice  & 80.69 & 79.95 & 80.99 &  81.32 & 81.67 & 81.51  & 81.90 & \textbf{82.51} \\
        3D Dice  & 87.40 & 86.65 & 88.07 &  88.24 & 88.48 & 88.48 & 88.50 & \textbf{89.06}\\
    \bottomrule
    \end{tabular}}
    \label{tab:results}
\end{table}
\vspace{-1em}

Table \ref{tab:results} gathers the model's segmentation performance after 10 cycles in terms of mean 2D Dice and mean 3D Dice scores over whole test volumes. In the fully-supervised setting, test-time augmentation-based sampling (TTA) outperforms random sampling, core-set selection, entropy sampling and sampling based on dropout. Similarly, unweighted TAAL and TAAL outperform random sampling in both semi-supervised and fully-supervised settings. After labeling 10 extra samples, the mean 3D Dice score attains 89.06\% with TAAL while only reaching respectively 87.40\% and 88.48\% with random sampling in fully- and semi-supervised settings. Similar results were observed with 2D Dice on test images.

\section{Conclusion}
In this paper, we presented a simple, yet effective semi-supervised deep active learning approach for medical image segmentation.  Our method, Test-time Augmentation for Active Learning (TAAL), employs a cross-augmentation consistency framework that produces both an improved training due to its unsupervised consistency loss, and a better sampling method through the uncertainty measure it provides. TAAL also uses a modified JSD that significantly improves the model's performance. Our results on the ACDC cardiac segmentation dataset show that, with TAAL, the trained model can reach up to 89.06\% 3D Dice with 20 labeled samples when it only reaches 87.40\% with random sampling. 
Because our approach exploits standard augmentation techniques already used in medical image segmentation tasks, TAAL emerges as a simple, yet efficient semi-supervised active learning strategy. 
While our method highly depends on the presence of disagreeing predictions for augmented inputs to identify the most informative samples, our observed improvements on a cardiac MRI dataset highlight promising avenues for future work, notably the investigation of more complex datasets and types of augmentations.\\

\noindent\textbf{Acknowledgments} -- This work is supported by the Canada
Research Chair on Shape Analysis in Medical Imaging, and the
Research Council of Canada (NSERC). Computational resources were partially provided by Compute Canada. The authors also thank the ACDC Challenge organizers for providing the data.

\bibliographystyle{splncs}
\bibliography{main}

\end{document}